\documentclass[12pt]{article}

\usepackage[margin=1.3in]{geometry}
\usepackage{times}
\usepackage{latexsym}
\usepackage[T1]{fontenc}
\usepackage[utf8]{inputenc}
\usepackage{microtype}
\usepackage{inconsolata}
\usepackage{graphicx}
\usepackage{natbib}
\usepackage{url}
\usepackage{authblk}

\usepackage{algorithm}
\usepackage{algorithmic}
\usepackage{algorithm}
\usepackage{algorithmic}
\usepackage{amsfonts}
\usepackage{amsmath}
\usepackage{bm}

\usepackage{booktabs}
\usepackage{subcaption}
\usepackage{float}

\newcommand{\customsmall}{\fontsize{8}{10}\selectfont}

\title{\Large Language Models for Longitudinal Clinical Prediction}

\author[1*]{Tananun Songdechakraiwut}
\author[2+]{Michael Lutz}
\date{}
\affil[1]{\small Department of Computer Science, Duke University}
\affil[2]{\small Departments of Neurology and Pathology, Duke University School of Medicine}
\affil[*]{\small \texttt{t.song@duke.edu}}
\affil[+]{\small \texttt{michael.lutz@duke.edu}}

\begin{document}

\maketitle

\begin{abstract}
We explore a lightweight framework that adapts frozen large language models to analyze longitudinal clinical data. The approach integrates patient history and context within the language model space to generate accurate forecasts without model fine-tuning. Applied to neuropsychological assessments, it achieves accurate and reliable performance even with minimal training data, showing promise for early-stage Alzheimer's monitoring.
\end{abstract}

\section{Introduction}

Alzheimer's disease is a progressive neurodegenerative disorder that develops gradually over many years. Early detection and personalized monitoring are essential for improving patient outcomes, enabling timely interventions, and optimizing clinical trial design \citep{van2023lecanemab}. Longitudinal neuropsychological assessments collected over multiple visits offer valuable insight into cognitive decline \citep{albert2011diagnosis}. However, real-world clinical data is often sparse, irregular, and incomplete, making accurate forecasting of future cognitive outcomes an ongoing challenge.

Large language models (LLMs) have shown strong generalization and reasoning capabilities across diverse domains \citep{brown2020language}. These models can be adapted to structured temporal data without retraining \citep{jin2024timellm}. Inspired by this, we explore whether pretrained LLMs can forecast future neuropsychological scores based on longitudinal patient histories.

Our approach adapts a frozen LLM to model clinical trajectories without fine-tuning. Longitudinal patient records are segmented, embedded, and translated into token sequences interpretable by the LLM. Patient demographics and information about what the model should predict are integrated through prompt-based conditioning, enabling the model to reason toward clinical objectives. The LLM's output is used to estimate future health indicators or risk scores.

We evaluate the method on longitudinal cognitive and functional assessments, showing that it produces accurate predictions across these assessment scores, even when trained with minimal annotated data, with potential for real-world use in early-stage Alzheimer's, where decline may unfold unevenly across domains.

\section{Longitudinal LLM Framework}

\begin{figure}[t]
  \centering
  \includegraphics[width=.65\columnwidth]{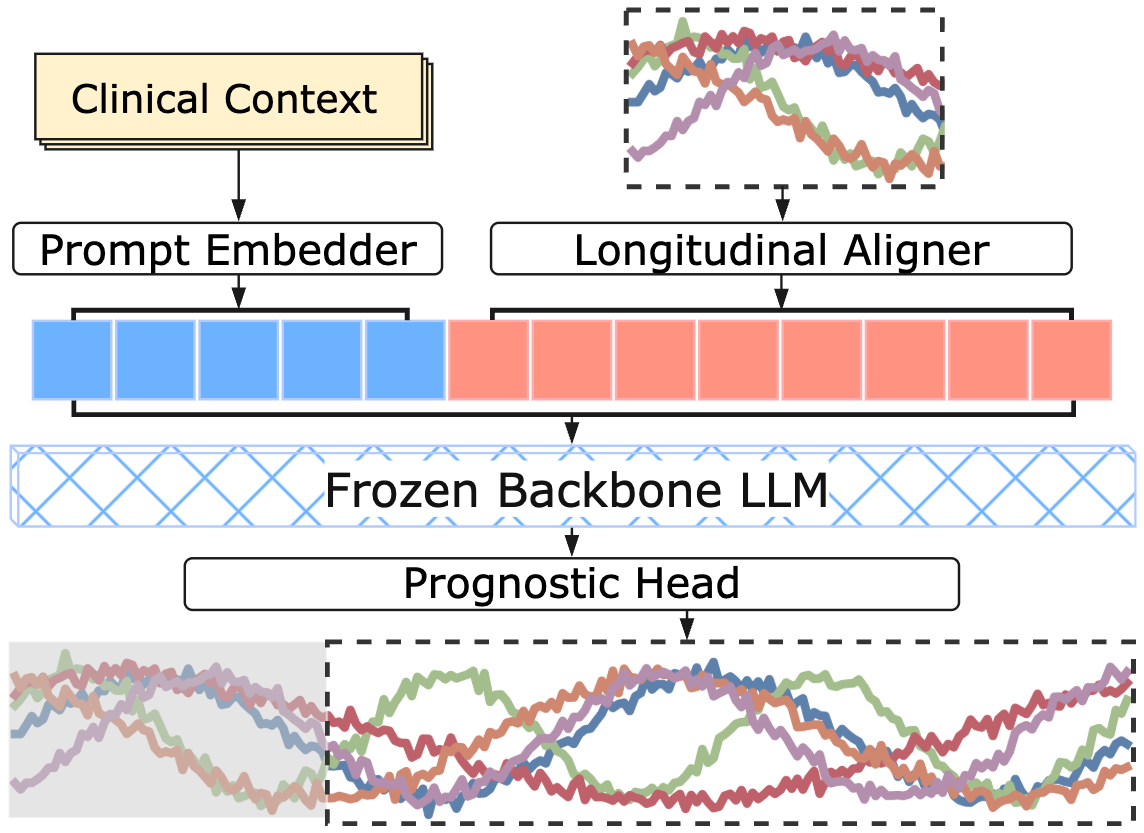}
  \caption{Schematic of a longitudinal LLM framework.}
  \label{fig:schematic}
\end{figure}

We investigate the use of a frozen LLM for forecasting clinical outcomes from longitudinal patient data. A schematic of our approach is shown in Figure~\ref{fig:schematic}. Each patient is represented by a combination of static clinical information (e.g., demographics) and a temporal sequence of clinical observations collected over multiple visits (e.g., neuropsychological assessments). The model takes these as input and predicts future values for selected clinical variables across a fixed forecasting horizon.
Formally, we denote the longitudinal data as a multivariate time series $\bm{X} \in \mathbb{R}^{d \times T}$, where $d$ is the number of variables and $T$ is the number of observed visits. The task is to forecast $\hat{\bm{Y}} \in \mathbb{R}^{d \times h}$, the values of those variables over the next $h$ time steps, with predictions aligned to future visits. Only the task-specific components are trained, while the LLM remains frozen.

\subsection{Prompt-based clinical conditioning}

We leverage the LLM to reason over longitudinal clinical data by framing the prediction task as a sequence modeling problem. To enable the LLM to interpret structured patient data, we prepend a textual prompt \citep{jin2024timellm} that encodes high-level clinical context and guides the model's attention during inference.

\begin{figure}[t]
  \includegraphics[width=\columnwidth]{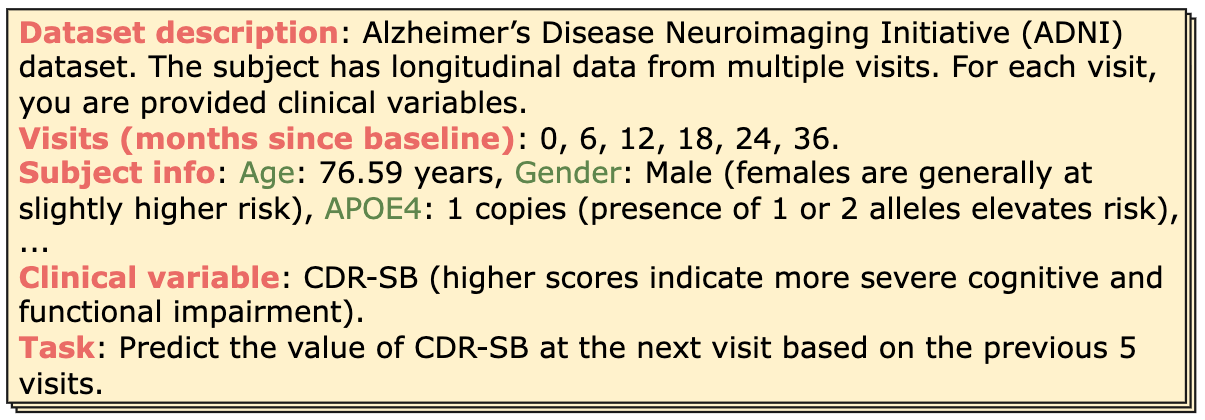}
  \caption{Clinical prompt example.}
  \label{fig:prompt_example}
\end{figure}

An example of the prompt format is illustrated in Figure~\ref{fig:prompt_example}. It typically includes relevant background information such as patient profile, task instruction, references to the clinical variables of interest, and a timeline indicating when visits occurred. The prompt is converted into token embeddings using the LLM's vocabulary and embedding matrix, resulting in a sequence $\bm{p} \in \mathbb{R}^{\ell_p \times d_h}$, where $\ell_p$ is the prompt length and $d_h$ is the hidden dimension of the model.
This prompt is later concatenated with tokenized representations of the patient's longitudinal measurements and the full sequence is processed jointly by the LLM. This setup allows the model to incorporate background knowledge and conditioning signals from the prompt as it reasons over the patient's longitudinal trajectory.

\subsection{Temporal embedding of clinical visits}

The longitudinal input $\bm{X}$ is decomposed into univariate sequences $\{\bm{X}^{(i)} \in \mathbb{R}^{1 \times T}\}_{i=1}^d$ and processed independently \citep{nie2022time}. Each sequence is normalized using reversible~instance~normalization (RevIN) \citep{kim2021reversible}, which centers the data to zero mean and unit variance. This mitigates variability in scale and baseline that commonly arises across patients due to demographic or diagnostic differences, allowing the model to focus on relative temporal changes (e.g., cognitive decline) rather than absolute score levels. The transformation is reversible, enabling predictions to be mapped back to the original clinical scale during inference.

To capture local temporal structure, each normalized sequence is segmented into overlapping fixed-length windows, or temporal patches \citep{nie2022time}, using a sliding window of length $\ell$ and stride $s$:
$\Tilde{\bm{X}}^{(i)} \in \mathbb{R}^{m \times \ell}, \text{where } m = \left\lfloor \frac{T - \ell}{s} \right\rfloor + 2$.
In real-world clinical data, visit intervals are often irregular and measurements may be missing due to skipped visits or incomplete assessments. These temporal inconsistencies can be addressed through strategies such as imputation, masking, or interpolation to a reference timeline. Additionally, encoding the time between visits using elapsed time, time embeddings, or attention-based mechanisms allows models such as transformers \citep{li2020behrt} to account for variable intervals and more accurately capture disease progression dynamics.

Each patch is then mapped into a fixed-dimensional vector using a learnable transformation (e.g., a linear projection or a small multi-layer perceptron), resulting in an embedded patch sequence $\hat{\bm{X}}^{(i)} \in \mathbb{R}^{m \times d_e}$, where $d_e$ is the patch embedding dimension. These patch embeddings form a sequence of compact token representations that summarize localized progression patterns across adjacent visits and are compatible with the token structure expected by the LLM.

\subsection{Semantic mapping of temporal patterns}

Since large language models are trained to operate on natural language tokens, the numerical patch embeddings from clinical time series must be aligned with the model's input space. To achieve this, we employ a cross-modal reprogramming mechanism that maps each patch embedding into the LLM's token space using a compact set of semantic anchors~\citep{jin2024timellm}.

Specifically, each input patch sequence $\hat{\bm{X}}^{(i)} \in \mathbb{R}^{m \times d_e}$ is transformed into a reprogrammed sequence $\bm{z}^{(i)} \in \mathbb{R}^{m \times d_h}$, where $d_h$ is the hidden dimension of the frozen language model. This transformation is implemented via \emph{multi-head} cross-attention between the patch tokens and a fixed set of \emph{text prototypes}, defined as a subset of token embeddings selected from the LLM's vocabulary. These prototypes serve as a semantic basis for aligning clinical inputs to the language model's representation space.
Multiple strategies may be used to refine the prototype set. Semantic filtering can prioritize clinically relevant descriptors based on embedding similarity to medical keywords or text-mined vocabulary from domain corpora. Alternatively, clustering the vocabulary embedding space or applying sparsity constraints in attention may yield more diverse or compact sets. Learnable prototypes or reparameterized projection layers may also be used to adapt the prototype space during training.

These prototypes act as semantic intermediaries that allow the model to interpret temporal patterns in clinically meaningful terms. For example, distinct tokens may implicitly represent archetypal changes like ``sharp decline,'' ``gradual improvement,'' or ``stable trajectory.'' The resulting reprogrammed output $\bm{z}^{(i)}$ is passed to the language model alongside the clinical prompt, enabling reasoning over structured patient timelines within the LLM's native input space.

\subsection{LLM integration and forecasting}

To prepare the full input for the LLM, we concatenate the clinical prompt $\bm{p}$ with the reprogrammed patch embeddings $\bm{z}^{(i)}$ for each clinical variable. This produces a unified token sequence $\bm{X}^{(i)}_{\text{LLM}} \in \mathbb{R}^{(\ell_p + m) \times d_h}$. The prompt provides high-level context (e.g., patient profile, task specification), while the reprogrammed patch tokens encode localized temporal patterns from the patient's visit history.

The full sequence $\bm{X}^{(i)}_{\text{LLM}}$ is passed to the frozen LLM, which generates contextualized token-level representations $\mathbf{h}^{(i)} \in \mathbb{R}^{(\ell_p + m) \times d_h}$. 
These outputs capture interactions between the prompt and the temporal embeddings, enabling the model to reason over longitudinal cognitive trajectories in a semantically informed space.
They are then passed to a lightweight prediction module, which maps them to the desired forecasting targets. This module may take the form of a projection head, feedforward layer, or task-specific decoder, depending on the prediction objective. The output for each clinical variable is $\hat{\bm{Y}}^{(i)} \in \mathbb{R}^{1 \times h}$, where $h$ is the forecasting horizon.
If input normalization was applied during preprocessing (e.g., using RevIN), a corresponding denormalization step is applied to the outputs, restoring predictions to their original clinical scale. The final forecast $\hat{\bm{Y}} \in \mathbb{R}^{d \times h}$ represents a multivariate trajectory aligned with expected neuropsychological assessments over future visits, and may be used for evaluation, risk stratification, or clinical decision support.

\subsection{Efficient training and clinical adaptability}

The model is trained using mean squared error between predicted and ground-truth neuropsychological scores across all variables and forecasted time steps:
$\mathcal{L}_{\mathrm{MSE}} = \frac{1}{d h} \sum_{i,j=1}^{d,h} ( \hat{Y}^{(i)}_j - Y^{(i)}_j )^2$.

Importantly, only a small subset of components are updated during training: the patch embedding layer, cross-modal reprogramming module, and the forecasting head. The backbone language model remains frozen. This setup enables efficient learning with minimal computational overhead and supports generalization in low-data regimes, which is particularly important in clinical contexts where annotated longitudinal data is often limited. In such settings, where patient cohorts are small, visit frequencies are irregular, and data collection is costly, the ability to learn from few examples is essential. 
The approach is also compatible with model compression like quantization and adapters, supporting flexible clinical deployment.

\begin{table*}
\customsmall
\centering

\setlength{\tabcolsep}{3pt}

\begin{subtable}{\textwidth}
\centering
\begin{tabular}{r|cccc|cccc}
\toprule
        & \multicolumn{4}{c|}{GPT-2} & \multicolumn{4}{c}{BERT} \\
Mo &   \texttt{FAQTOTAL} &   \texttt{AVDEL30} &   \texttt{ADAS13} &   \texttt{CDR-SB} &  
 \texttt{FAQTOTAL} &   \texttt{AVDEL30} &   \texttt{ADAS13} &   \texttt{CDR-SB} \\
\midrule
12 & 1.831\,$\pm$\,0.098 & 1.677\,$\pm$\,0.104 & 3.538\,$\pm$\,0.063 & 0.749\,$\pm$\,0.057 & 1.823\,$\pm$\,0.104 & 1.680\,$\pm$\,0.097 & 3.543\,$\pm$\,0.094 & 0.749\,$\pm$\,0.057 \\
18 & 2.361\,$\pm$\,0.481 & 1.412\,$\pm$\,0.174 & 3.851\,$\pm$\,0.492 & 0.925\,$\pm$\,0.207 & 2.333\,$\pm$\,0.472 & 1.448\,$\pm$\,0.168 & 3.899\,$\pm$\,0.485 & 0.922\,$\pm$\,0.194 \\
24 & 1.799\,$\pm$\,0.126 & 1.732\,$\pm$\,0.067 & 3.551\,$\pm$\,0.241 & 0.732\,$\pm$\,0.077 & 1.802\,$\pm$\,0.121 & 1.726\,$\pm$\,0.051 & 3.626\,$\pm$\,0.277 & 0.739\,$\pm$\,0.080 \\
36 & 2.158\,$\pm$\,0.232 & 1.962\,$\pm$\,0.068 & 3.769\,$\pm$\,0.216 & 0.788\,$\pm$\,0.058 & 2.181\,$\pm$\,0.220 & 1.969\,$\pm$\,0.086 & 3.743\,$\pm$\,0.167 & 0.797\,$\pm$\,0.065 \\
48 & 1.726\,$\pm$\,0.079 & 2.065\,$\pm$\,0.229 & 3.409\,$\pm$\,0.269 & 0.645\,$\pm$\,0.058 & 1.706\,$\pm$\,0.191 & 2.171\,$\pm$\,0.206 & 3.473\,$\pm$\,0.297 & 0.672\,$\pm$\,0.038 \\
\bottomrule
\end{tabular}
\caption{70\% few-shot learning}
\label{tab:mae_all_train_70}
\end{subtable}

\vspace{0.11em}

\begin{subtable}{\textwidth}
\centering
\begin{tabular}{r|cccc|cccc}
\toprule
        & \multicolumn{4}{c|}{GPT-2} & \multicolumn{4}{c}{BERT} \\
Mo &   \texttt{FAQTOTAL} &   \texttt{AVDEL30} &   \texttt{ADAS13} &   \texttt{CDR-SB} &  
 \texttt{FAQTOTAL} &   \texttt{AVDEL30} &   \texttt{ADAS13} &   \texttt{CDR-SB} \\
\midrule
12 & 1.824\,$\pm$\,0.028 & 1.655\,$\pm$\,0.046 & 3.455\,$\pm$\,0.054 & 0.741\,$\pm$\,0.014 & 1.799\,$\pm$\,0.032 & 1.654\,$\pm$\,0.037 & 3.448\,$\pm$\,0.051 & 0.740\,$\pm$\,0.014 \\
18 & 2.399\,$\pm$\,0.149 & 1.350\,$\pm$\,0.100 & 3.765\,$\pm$\,0.175 & 0.806\,$\pm$\,0.033 & 2.460\,$\pm$\,0.143 & 1.353\,$\pm$\,0.129 & 3.769\,$\pm$\,0.133 & 0.809\,$\pm$\,0.041 \\
24 & 1.894\,$\pm$\,0.066 & 1.812\,$\pm$\,0.109 & 3.714\,$\pm$\,0.048 & 0.763\,$\pm$\,0.033 & 1.838\,$\pm$\,0.042 & 1.851\,$\pm$\,0.106 & 3.661\,$\pm$\,0.118 & 0.766\,$\pm$\,0.028 \\
36 & 2.143\,$\pm$\,0.071 & 2.037\,$\pm$\,0.089 & 3.881\,$\pm$\,0.085 & 0.787\,$\pm$\,0.017 & 2.151\,$\pm$\,0.042 & 1.932\,$\pm$\,0.046 & 3.893\,$\pm$\,0.091 & 0.799\,$\pm$\,0.022 \\
48 & 1.768\,$\pm$\,0.068 & 2.233\,$\pm$\,0.143 & 3.642\,$\pm$\,0.084 & 0.698\,$\pm$\,0.023 & 1.783\,$\pm$\,0.027 & 2.283\,$\pm$\,0.107 & 3.606\,$\pm$\,0.105 & 0.707\,$\pm$\,0.035 \\
\bottomrule
\end{tabular}
\caption{10\% few-shot learning}
\label{tab:mae_all_train_10}
\end{subtable}

\vspace{0.11em}

\begin{subtable}{\textwidth}
\centering
\begin{tabular}{r|cccc|cccc}
\toprule
        & \multicolumn{4}{c|}{GPT-2} & \multicolumn{4}{c}{BERT} \\
Mo &   \texttt{FAQTOTAL} &   \texttt{AVDEL30} &   \texttt{ADAS13} &   \texttt{CDR-SB} &  
 \texttt{FAQTOTAL} &   \texttt{AVDEL30} &   \texttt{ADAS13} &   \texttt{CDR-SB} \\
\midrule
12 & 1.890\,$\pm$\,0.048 & 1.660\,$\pm$\,0.034 & 3.479\,$\pm$\,0.031 & 0.737\,$\pm$\,0.017 & 1.849\,$\pm$\,0.033 & 1.675\,$\pm$\,0.039 & 3.482\,$\pm$\,0.064 & 0.737\,$\pm$\,0.017 \\
18 & 2.749\,$\pm$\,0.275 & 1.201\,$\pm$\,0.061 & 3.995\,$\pm$\,0.272 & 0.894\,$\pm$\,0.031 & 2.955\,$\pm$\,0.250 & 1.204\,$\pm$\,0.100 & 4.143\,$\pm$\,0.316 & 0.910\,$\pm$\,0.054 \\
24 & 2.146\,$\pm$\,0.142 & 1.760\,$\pm$\,0.026 & 3.837\,$\pm$\,0.155 & 0.845\,$\pm$\,0.023 & 2.015\,$\pm$\,0.138 & 1.865\,$\pm$\,0.105 & 3.758\,$\pm$\,0.032 & 0.806\,$\pm$\,0.044 \\
36 & 2.566\,$\pm$\,0.088 & 1.861\,$\pm$\,0.225 & 3.996\,$\pm$\,0.033 & 0.916\,$\pm$\,0.030 & 2.438\,$\pm$\,0.127 & 2.159\,$\pm$\,0.264 & 4.068\,$\pm$\,0.063 & 0.878\,$\pm$\,0.037 \\
48 & 2.195\,$\pm$\,0.196 & 2.107\,$\pm$\,0.198 & 4.119\,$\pm$\,0.094 & 0.869\,$\pm$\,0.066 & 2.128\,$\pm$\,0.204 & 2.276\,$\pm$\,0.271 & 4.051\,$\pm$\,0.086 & 0.833\,$\pm$\,0.053 \\
\bottomrule
\end{tabular}
\caption{1\% few-shot learning}
\label{tab:mae_all_train_1}
\end{subtable}

\caption{MAE\,$\pm$\,std for clinical forecasts over 12--48 month horizons under 70\%, 10\%, and 1\% training data settings. For each setting, our method is \emph{trained once} to jointly predict \emph{all} targets. Lower indicate better performance.}
\label{tab:mae_all_train_sizes}
\end{table*}

\begin{table}
\customsmall
\centering
\setlength{\tabcolsep}{3.5pt}
\begin{tabular}{r|cccccc}
\toprule
Mo & Default & 6L & No Prompt & No RevIN & 1000P & 4H \\
\midrule
12 & 0.7406 & 0.7403 & 0.7404 & 2.2433 & 0.7407 & 0.7405 \\
18 & 0.8060 & 0.8344 & 0.8217 & 5.0340 & 0.8053 & 0.8045 \\
24 & 0.7627 & 0.7803 & 0.7737 & 2.5498 & 0.7577 & 0.7594 \\
36 & 0.7873 & 0.7903 & 0.8235 & 3.4318 & 0.8024 & 0.7899 \\
48 & 0.6978 & 0.7398 & 0.7368 & 2.8905 & 0.7060 & 0.7052 \\
\bottomrule
\end{tabular}
\caption{Ablation study on \texttt{CDR-SB} forecasting MAE under 10\% training. Default: GPT-2, 12 LLM layers (L), 100 text prototypes (P), 8 attention heads (H), with prompt and RevIN.}
\label{tab:ablation_cdrsb}
\end{table}

\section{Forecasting Alzheimer's Progression}

We evaluate our approach on the Alzheimer's Disease Neuroimaging Initiative (ADNI) dataset \citep{jack2008alzheimer}, which includes longitudinal clinical data from 1,783 participants, each with up to 7 visits: baseline and follow-ups at approximately 6, 12, 18, 24, 36, and 48 months. Neuropsychological assessments are recorded at each visit when available; however, not all assessments are consistently recorded across timepoints due to incomplete follow-up.
We focus on four clinically relevant targets: \texttt{FAQTOTAL} (functional status) \citep{pfeffer1982measurement}, \texttt{AVDEL30} (30-minute delayed recall) \citep{crane2012development}, \texttt{ADAS13} (global cognition) \citep{mohs1997development}, and \texttt{CDR-SB} (disease staging) \citep{morris1993clinical}. These complementary measures are sensitive to early Alzheimer's, where prediction supports timely intervention \citep{albert2011diagnosis}. Relevant demographic features, including age, gender, APOE4 allele copies, and years of education, are incorporated via the clinical prompt.

We evaluate five prediction settings corresponding to 12, 18, 24, 36, and 48-month horizons, each using all available past visits to forecast the next follow-up. This simulates a realistic clinical scenario aimed at early diagnostic support. The dataset is split by subject into 70\% training, 10\% validation, and 20\% test sets. To assess generalization under limited supervision, we also evaluate models using 10\% and 1\% of the training data. All experiments are repeated across 5 independent random splits, and results are averaged. We evaluate two frozen language model backbones, GPT-2 \citep{radford2019language} and BERT \citep{devlin2019bert}, and report mean absolute error (MAE) across all forecasted variables and timepoints.

Full procedure, model configuration, and dataset details are provided in Appendices~\ref{app:llm_algorithm}, \ref{appx:model_config}, and \ref{appx:adni_data}.

\subsection{Results and clinical interpretation}

Table~\ref{tab:mae_all_train_70} reports the MAE of forecasted clinical outcomes using 70\% of the data for training. Both models show strong and consistent performance across timepoints and targets, with comparable accuracy in most cases. These results are clinically acceptable when compared to established minimal clinically important difference (MCID) thresholds \citep{muir2024minimal}: changes of 3--5 points on \texttt{FAQTOTAL}, 2--3 points on \texttt{ADAS13}, and 1--2 points on \texttt{CDR-SB} are generally considered meaningful decline, while \texttt{AVDEL30} scores typically vary by 1--2 words on its 0--15 scale. The model's forecast errors fall within these ranges, supporting its potential for early detection, longitudinal monitoring, and clinically actionable decision-making.

To assess generalization under limited supervision, we evaluate model performance when trained on only 10\% and 1\% of the original training set (Tables~\ref{tab:mae_all_train_10} and~\ref{tab:mae_all_train_1}). This reflects real-world clinical scenarios where longitudinal data with complete annotations is often scarce. With 10\% of the data, both models maintain performance close to the full-data setting, with only modest increases in MAE across variables and timepoints. Notably in the 1\% setting, \texttt{CDR-SB}, which is widely used to quantify disease severity and guide staging in both research and clinical care, remains consistently well predicted. Forecast errors for \texttt{CDR-SB} are well below the commonly accepted 1.0-point MCID threshold across all settings, including the most data-constrained regime. These results highlight the framework's ability to support clinically meaningful forecasting, even when training data is highly limited.

\paragraph{Ablation study.} Table~\ref{tab:ablation_cdrsb} shows that the default setup performs best overall. Removing RevIN causes large MAE increases, confirming its importance. Omitting the prompt results in higher errors. Reducing LLM layers, attention heads, or increasing text prototypes degrades long-horizon accuracy. Each component contributes incrementally, with RevIN and prompting being most critical.

\section*{Limitations}

While our approach demonstrates strong predictive performance, one limitation is the current level of interpretability. Although prompts are used to provide clinical context and guide the model's reasoning, the internal mapping between time series tokens and language model predictions remains somewhat opaque, which may present challenges for adoption in high-stakes medical settings where understanding the basis for predictions is important. Addressing this limitation is a promising direction for future work, including the integration of model explanation methods such as attention visualization, prototype attribution, or post-hoc interpretability techniques. Collaborating with clinical experts to qualitatively assess model outputs and align explanations with domain understanding could further enhance trust and practical utility.

\bibliographystyle{abbrvnat}
\bibliography{reference}

\clearpage
\appendix

\section{Algorithmic Implementation of Longitudinal LLM}
\label{app:llm_algorithm}

We detail the full procedural steps of our method in Algorithm~\ref{alg:longitudinalllm}, which outlines how structured patient data is reprogrammed and processed by a frozen LLM for clinical time series forecasting.

\section{Model Configurations}
\label{appx:model_config}

We report the model configuration used for next-visit forecasting, where predictions are made based on observed clinical data up to a given time point. The full setup is summarized in Table~\ref{tab:model_config}. By default, we use a frozen language model backbone, either GPT-2 or BERT, with 12 layers operating at full capacity. Patch length $\ell$ is set to 2 with a stride $s$ of 1 to capture dynamics across consecutive clinical visits. The patch embeddings are projected into a $d_e = 16$ dimensional space, and 8 attention heads are used in the semantic mapping of temporal patterns. The number of text prototypes $V'$ is fixed at 100 across all experiments. We train the models using the Adam optimizer~\citep{KingmaB14adam} with an initial learning rate of 0.005 and mean squared error (MSE) loss. Unless otherwise specified, these settings serve as our standard configuration.

\begin{table}[h]
\centering
\setlength{\tabcolsep}{6pt}
\begin{tabular}{l r}
\toprule
\textbf{Configuration} & \textbf{Value} \\
\midrule
Text Prototypes $V'$ & 100 \\
LLM Layers & 12 \\
Patch Length $\ell$ & 2 \\
Stride $s$ & 1 \\
Patch Embedding Dim.\ $d_e$ & 16 \\
Heads $K$ & 8 \\
Initial Learning Rate & $0.005$ \\
Loss & MSE \\
Epochs & 30 \\
\bottomrule
\end{tabular}
\caption{Default model configuration.}
\label{tab:model_config}
\end{table}

\paragraph{Experimental environment.}
The system environment included \textsc{Ubuntu} 24.04.2 LTS, an \textsc{AMD Ryzen Threadripper 7960X} (24 cores), 64\,GB of RAM, and 1\,TB NVMe SSD. It was equipped with an \textsc{NVIDIA GeForce RTX 4090} GPU, \textsc{NVIDIA} Driver 550.144.03, and \textsc{CUDA} 12.4.

\section{ADNI Dataset and Preprocessing}
\label{appx:adni_data}

We use the Alzheimer's Disease Neuroimaging Initiative (ADNI) dataset \citep{jack2008alzheimer} and preprocess it into a structured, visit-aligned format for longitudinal modeling. The original data is provided in tabular format and includes clinical assessments collected longitudinally across multiple visits. The dataset was downloaded on June 13, 2025 from the ADNI data repository.

\paragraph{Longitudinal features.}
We focus on seven canonical visit timepoints: baseline (\texttt{bl}), and follow-ups at 6, 12, 18, 24, 36, and 48 months.
From each visit, we extract the following variables: \texttt{CDRSB} (Clinical Dementia Rating Sum of Boxes) \citep{morris1993clinical}, \texttt{TOTAL13} (ADAS13 cognitive score) \citep{mohs1997development}, \texttt{FAQTOTAL} (Functional Activities Questionnaire total score) \citep{pfeffer1982measurement}, and \texttt{AVDEL30MIN} (30-minute delayed recall score) \citep{crane2012development}. These variables are commonly used in Alzheimer's disease research to monitor cognitive and functional changes. All values are converted to numeric format, with invalid or empty entries treated as missing. No imputation is performed. During model training, missing values are masked and do not contribute to the loss. Similarly, evaluation metrics are computed only over observed target values, ensuring that performance is assessed strictly on available ground truth.

\begin{table}
\centering
\begin{tabular}{l r}
\toprule
Variable & Value \\
\midrule
Age (years) & 73.5 $\pm$ 7.2 \\
Sex (female) & 811 (45.5\%) \\
APOE4 = 0 & 947 (53.1\%) \\
APOE4 = 1 & 656 (36.8\%) \\
APOE4 = 2 & 178 (10.0\%) \\
APOE missing & 2 (0.1\%) \\
Years of education & 16.0 $\pm$ 2.8 \\
Mean visits per participant & 4.8 $\pm$ 1.3 \\
\bottomrule
\end{tabular}
\caption{Summary statistics of the ADNI cohort used in this study. Values reflect the processed cohort after applying selection criteria. Visit count includes only visits with at least one observed score among \texttt{CDRSB}, \texttt{FAQTOTAL}, \texttt{AVDEL30MIN}, or \texttt{TOTAL13}.}
\label{tab:adni-stats}
\end{table}

\paragraph{Demographic features.}
We extract static features from the screening visit (\texttt{sc}): age at baseline (in years), sex (female or male), APOE $\varepsilon$4 allele count (0, 1, or 2), and years of education. These variables are commonly associated with Alzheimer's disease risk and progression, and are included in the model prompts.

\paragraph{Cohort selection.}
We include subjects with at least two non-missing values for \texttt{CDRSB} across the seven scheduled timepoints. Subjects with six or more missing \texttt{CDRSB} values are excluded. Other target variables may be missing due to incomplete follow-up or variability in data collection. No diagnosis-based filtering is applied.

\paragraph{Population summary.}
Table~\ref{tab:adni-stats} reports descriptive statistics for the final study cohort (\textbf{N = 1,783}). Continuous variables are summarized using the mean and standard deviation, and categorical variables are reported as counts and percentages.

\paragraph{Data source.}
Data used in the preparation of this article were obtained from the ADNI database (\url{adni.loni.usc.edu}). The ADNI was launched in 2003 as a public-private partnership, led by Principal Investigator Michael W. Weiner, MD. The primary goal of ADNI has been to test whether serial magnetic resonance imaging (MRI), positron emission tomography (PET), other biological markers, and clinical and neuropsychological assessment can be combined to measure the progression of mild cognitive impairment and early Alzheimer's disease. For up-to-date information, see \url{www.adni-info.org}.

\begin{algorithm*}
\caption{Longitudinal Forecasting with Frozen LLM}
\label{alg:longitudinalllm}
\begin{algorithmic}[1]
\REQUIRE Longitudinal time series $\bm{X} \in \mathbb{R}^{d \times T}$, frozen LLM $f(\cdot)$, forecast horizon $h$
\ENSURE Forecasted outputs $\hat{\bm{Y}} \in \mathbb{R}^{d \times h}$

\STATE \textbf{// Prompt-Based Clinical Conditioning}
\STATE Construct prompt text from patient profile and summary statistics of $\bm{X}$
\STATE Tokenize and embed prompt: $\bm{p} \in \mathbb{R}^{\ell_p \times d_h}$

\STATE \textbf{// Text Prototype Initialization}
\STATE Select or learn semantic prototypes: $\bm{E}' \in \mathbb{R}^{V' \times d_h}$ from LLM embedding matrix

\FOR{each clinical variable $i = 1$ to $d$}
    \STATE \textbf{// Temporal Embedding of Clinical Visits}
    \STATE Normalize univariate series $\bm{X}^{(i)} \in \mathbb{R}^{1 \times T}$ using RevIN
    \STATE Segment into $m$ temporal patches: $\tilde{\bm{X}}^{(i)} \in \mathbb{R}^{m \times \ell}$
    \STATE Embed patches: $\hat{\bm{X}}^{(i)} \in \mathbb{R}^{m \times d_e}$

    \STATE \textbf{// Semantic Mapping of Temporal Patterns}
    \FOR{each head $k = 1$ to $K$}
        \STATE Compute queries: $\bm{Q}_k^{(i)} = \hat{\bm{X}}^{(i)} \bm{W}^Q_k$
        \STATE Compute keys and values from text prototypes: $\bm{K}_k = \bm{E}' \bm{W}^K_k,\quad \bm{V}_k = \bm{E}' \bm{W}^V_k$
        \STATE Cross-attend: $\bm{Z}_k^{(i)} = \mathrm{Softmax}\left(\frac{\bm{Q}_k^{(i)} {\bm{K}_k}^\top}{\sqrt{d}}\right)\bm{V}_k$
    \ENDFOR
    \STATE Concatenate and project: $\bm{z}^{(i)} = \mathrm{Concat}(\bm{Z}_1^{(i)}, ..., \bm{Z}_K^{(i)}) \bm{W}^O + \bm{b}^O$

    \STATE \textbf{// LLM Integration and Forecasting}
    \STATE Form LLM input sequence: $\bm{X}_{\text{LLM}}^{(i)} = [\bm{p};\ \bm{z}^{(i)}]$
    \STATE Run LLM: $\bm{h}^{(i)} = f(\bm{X}_{\text{LLM}}^{(i)})$
    \STATE Extract patch outputs: $\bm{h}_{\text{patch}}^{(i)} = \bm{h}^{(i)}[\ell_p:]$

    \STATE \textbf{// Forecast Projection}
    \STATE Map to forecast: $\hat{\bm{Y}}^{(i)} = \mathrm{PredictionHead}(\bm{h}_{\text{patch}}^{(i)})$
    \STATE Apply denormalization if used
\ENDFOR

\RETURN $\hat{\bm{Y}} = [\hat{\bm{Y}}^{(1)}, ..., \hat{\bm{Y}}^{(d)}]^\top$
\end{algorithmic}
\end{algorithm*}

\end{document}